\documentclass[letterpaper, 10 pt, conference]{ieeeconf}  

\IEEEoverridecommandlockouts                              

\usepackage{tikz}
\usetikzlibrary{positioning, fit, calc, chains, matrix}

\usepackage[ruled,vlined]{algorithm2e}
\usepackage[pagebackref=true,breaklinks=true,letterpaper=true,colorlinks,bookmarks=false]{hyperref}
\usepackage{subcaption}

\title{\LARGE \bf
Innovative Integration of Visual Foundation Model with a Robotic Arm on a Mobile Platform
}

\author{Shimian Zhang$^{1}$ 
\thanks{$^{1}$Shimian Zhang is with Shanghai Hrstek Co., Ltd
        {\tt\small zhangsm@hrstek-robotics.com}}%
Qiuhong Lu$^{2}$ 
\thanks{$^{2}$Qiuhong Lu is with Shanghai Hrstek Co., Ltd
        {\tt\small luqh@hrstek-robotics.com}}%
}

\begin{document}

\maketitle
\thispagestyle{empty}
\pagestyle{empty}

\begin{abstract}

In the rapidly advancing field of robotics, the fusion of state-of-the-art visual technologies with mobile robotic arms has emerged as a critical integration. 
This paper introduces a novel system that combines the Segment Anything model (SAM) — a transformer-based visual foundation model — with a robotic arm on a mobile platform. 
The design of integrating a depth camera on the robotic arm's end-effector ensures continuous object tracking, significantly mitigating environmental uncertainties.
By deploying on a mobile platform, our grasping system has an enhanced mobility, playing a key role in dynamic environments where adaptability are critical.
This synthesis enables dynamic object segmentation, tracking, and grasping.
It also elevates user interaction, allowing the robot to intuitively respond to various modalities such as clicks, drawings, or voice commands, beyond traditional robotic systems.
Empirical assessments in both simulated and real-world demonstrate the system's capabilities. 
This configuration opens avenues for wide-ranging applications, from industrial settings, agriculture, and household tasks, to specialized assignments and beyond.
\end{abstract}

\section{Introduction}
\begin{figure}[!t]
    \centering
    \includegraphics[width=\linewidth]{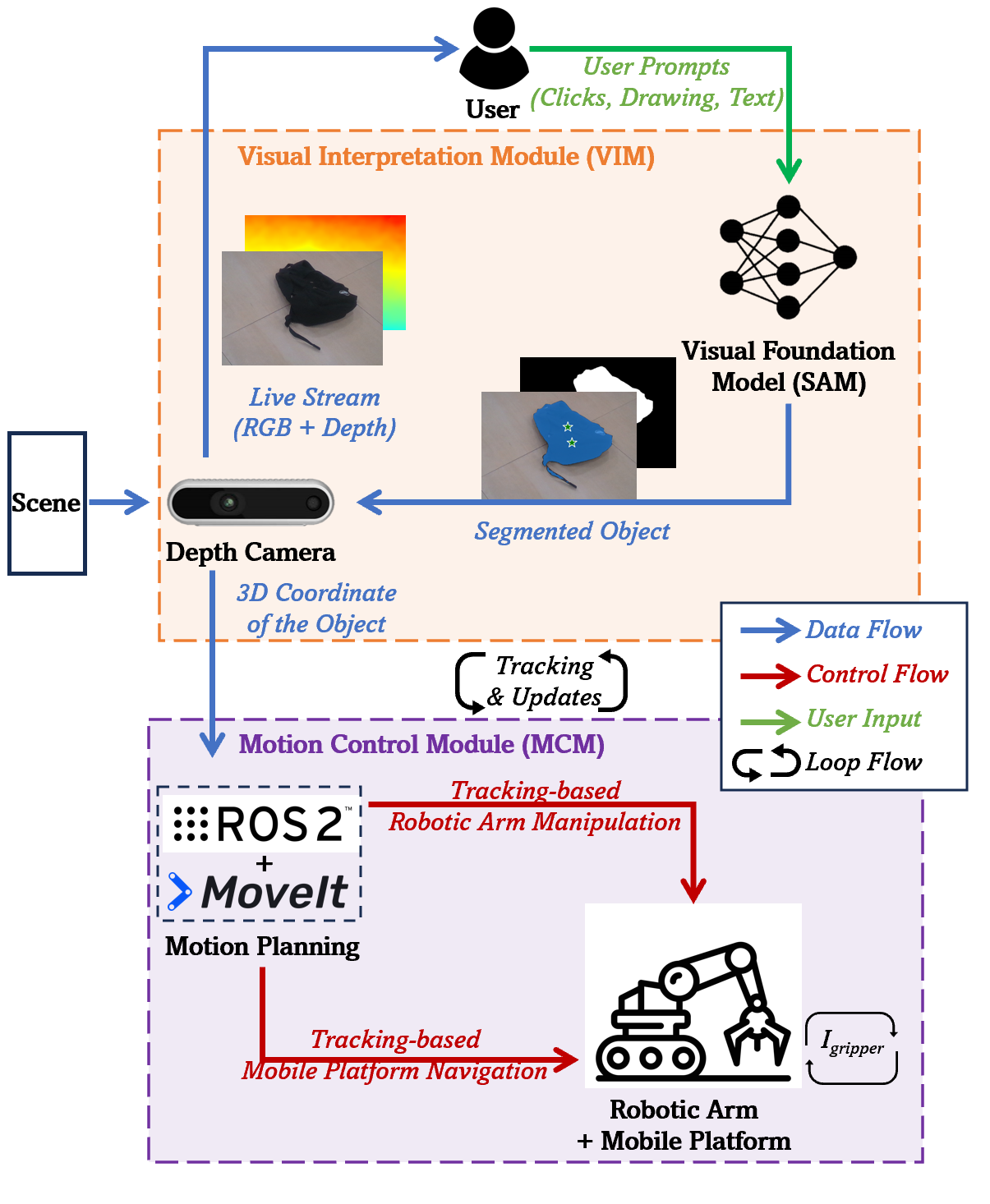}
    \caption{System Overview: Our integrated mobile robotic grasping system comprises two core modules: the Visual Interpretation Module (\textbf{VIM}) and the Motion Control Module (\textbf{MCM}). \textbf{VIM}, utilizing a depth camera, captures a live scene and, through the SAM visual foundation model, segments the user-indicated object for grasping. It then computes the object's 3D coordinates, relaying this data to \textbf{MCM}. Based on this localization, \textbf{MCM} plans the motion, determining if the platform needs relocation for optimal grasping. The robotic arm's movement leverages inverse kinematics for precision, with a closed-loop control anchored on continuous object tracking via the "eye-in-hand" system. The process ends with controlled grasping using force feedback from the arm's end effector.
    }
    \label{fig:robot-arm-diagram}
\end{figure}
The integration of vision systems into robotic arms has redefined the scope and efficiency of automation across numerous sectors \cite{du2021vision, huang2023voxposer, vemprala2023chatgpt}. These systems have become a cornerstone in industries where precision and repeatability are paramount. However, traditional vision-embedded robotic arms often grapple with challenges when faced with unrecognized objects, due to the limitation of object detection models \cite{jiang2022review, liu2016ssd}.
Their performance, though exceptional for familiar entities, deteriorates significantly with novel items. 
This constraint is further exacerbated by the inherent uncertainty in object detection performance, which often requires extensive fine-tuning.
Additionally, many of these systems are not capable to process and response instructions delivered in natural human language, which hinders their adaptability and user-friendliness.

In light of these challenges, we introduce a design that seamlessly integrates the state-of-the-art visual foundation model, the Segment Anything Model (SAM) \cite{kirillov2023segment}, into robotic arms. 
When combined with a mobile platform, this design has the potential to revolutionize robotic arm functionalities. Our innovative approach boasts a wide range of advantages:
\begin{itemize}
    \item \textbf{Universal Object Recognition:} The system's adeptness at segmenting and identifying a diverse set of objects negates the need for continual retraining and minimizes associated costs.
    \item \textbf{Enhanced Human Interaction:} Operators are presented with a spectrum of interaction modalities, ranging from clicks and spatial drawings to natural language prompts, fostering an intuitive user experience.
    \item \textbf{"Eye-in-Hand" System:} With the vision system integral to the arm's design, precise closed-loop control is achieved. Real-time object tracking further augments this, ensuring continuous object localization and optimal grasping.
    \item \textbf{Mobile Platform Integration:} By housing the robotic arm on a mobile platform, its operational scope is significantly broadened. Vision techniques augment traditional tasks and furnish a comprehensive understanding of dynamic environments, facilitating both enhanced navigation and strategic grasping.
\end{itemize}

The versatility of our mobile robotic arm design paves the way for applications spanning multiple domains, not limited to:
\begin{itemize}
    \item \textbf{Industrial Manufacturing:} It promises to revolutionize processes by adeptly identifying and managing a myriad of components, thus eliminating the requirement for continual system retraining. Further, the integrated mobile platform magnifies the robot's operational breadth, marking it as an indispensable tool within expansive industrial terrains.
    
    \item \textbf{Consumer Environments:} This design's potential materializes prominently in public and service sectors. Picture these robotic entities operating in settings like grocery stores, restaurants, and hotels, fulfilling tasks based on directives from customers or staff.
    
    \item \textbf{Specialized Scenarios:} Where our design truly stands unparalleled is in high-stakes, critical situations, such as taking care of disables, risky item removal. The blend of precision and adaptability, coupled with the enhanced coverage of the mobile platform, makes this system a choice for diverse, especially high-risk environments.
\end{itemize}


\section{Related Works}
\subsection{Visual Foundation Models}
A paradigm in the field of visual foundation models is the Segment Anything Model (SAM) by Meta AI Research \cite{kirillov2023segment}. SAM has emerged as a vision foundation model \cite{zhang2023survey} recognized for its unparalleled proficiency in segmenting any object within an image. Characterized by its transformer-based structure, SAM is trained on a large segmentation dataset with over a billion masks. 
Different from the other segmentation models, SAM performs promptable segmentation without the reliance on labeled regions. Instead, it operates on varied prompts, ranging from point prompts via clicks, region prompts through drawing, to text prompts in natural language. This versatility is underscored by its impressive zero-shot transfer performance, obviating any requirement for finetuning. Furthermore, the model's adaptability has catalyzed a myriad of applications across diverse scenarios like scene understanding \cite{chen2023bridging}, inpainting \cite{yu2023inpaint}, object tracking \cite{cheng2023segment}, and pose estimation \cite{fan2023pope}. 

In this work, we introduce a robotic grasping system based on SAM, envisioning to improve the quality of traditional robotic systems through this cutting-edge vision technology. 
At the same time, our research also includes an evaluation of SAM derivatives, especially Mobile SAM \cite{zhang2023faster}. 
This streamlined variant, achieved through knowledge distillation, results in a model that is approximately 60 times more compact than its predecessor, while still performing on par with the original SAM.

\subsection{Language-driven Robotic Manipulation}
With the emergence of large language models (LLMs), research of embedding language-driven technology into robot operations has been proposed \cite{huang2023voxposer, zhang2023bridging, vemprala2023chatgpt}.
\cite{huang2023voxposer} recently introduced an innovative robotic manipulation approach named VoxPoser. 
This method seamlessly integrates a large language model (LLM) with a visual-language model (VLM). Central to their approach is the LLM's capability to interpret free-form language instructions, which in tandem with its code-generation capacity, interfaces with the VLM. 
This collaborative interaction translates 3D value maps into spatially grounded knowledge to the agent, which is subsequently utilized in model-based planning frameworks to guide the agent's movements. 

While their methodology differs from ours, both approaches utilize foundation models to enhance traditional robotic systems, making them more responsive to human natural language directives.
Nevertheless, their system is not without its challenges. 
However, their system has notable limitations. Its environmental understanding is constrained, particularly when representing complex objects for grasping, such as irregular packages, complex assemblies, etc.
Although the LLMs are good at natural language interpretation and code generation, they have a potential to issue erroneous commands due to its hallucination phenomena, as echoed in \cite{vemprala2023chatgpt, bubeck2023sparks, bang2023multitask}

Our work aims to bridge these identified gaps. It favors visual foundation models for a more representative environmental understanding and employs strategies with depth cameras, inverse kinematics, and motion planning, that already institutionalized in the industry, for robotic arm manipulation. Such a methodology significantly diminishes uncertainties common in LLMs.

\subsection{Integrating SAM into Robotic Grasping}
\cite{yang2023pave} recently explored the incorporation of SAM into robotic grasping paradigms, which presents a noteworthy advancement in robotic grasping.
However, it has several remarkable limitations. 
First, their design utilizes dual fixed cameras, in contrast to our “eye-in-hand” system. An "eye-in-hand" setup is more adapt to real-world scenarios, particularly when the robotic arm is stationed in the wild environment. In dynamic environments, fixed camera positions are often impractical, emphasizing the importance of a robot-oriented vision system. 
Second, their choice to compile a dataset for training grasping strategies somewhat deviates from the essence of using SAM as a ``zero-shot'' foundation model without the need for specialized training or fine-tuning. 
Relying on such a dataset potentially limits the system's adaptability across diverse real-world tasks and settings.

In contrast to their approach, we've integrated an "eye-in-hand" system with our robotic arm on a mobile platform, enhancing its adaptability in diverse environments. Bypassing the need for grasp strategy training, we utilize the Denavit-Hartenberg (D-H) configurations \cite{denavit1955kinematic}, inverse kinematic estimations and continuous object tracking for precise movements. Our methodology not only eliminates the need for task-specific training datasets, avoiding potential data biases, but also capitalizes on the power of foundational models, presenting a cost-effective solution for both industrial and consumer use.



\section{Methodology}
\label{sec:methodology}

\begin{algorithm}
\SetAlgoLined
\KwIn{Live video stream from the depth camera, User prompts for object of interest}


\tcp{Object Segmentation}
frame, depth = captureStream()\;
segment, mask = SAM(frame, user\_prompt)\;

\tcp{Depth Estimation \& Positioning}
$\bar{d}$ = avgDepth(mask, depth)\;
$\mathbf{P}_{cam}$ = computeCamCoords($\bar{d}$, mask)\;

$\mathbf{P}_{arm}$ = transformCoords($\mathbf{P}_{cam}$)\;

\tcp{Platform Movement}
\If{$\mathbf{P}_{arm}$ is out of reach}{
    movePlatform($\mathbf{P}_{arm}$)\;
}

\tcp{Inverse Kinematics}
$\mathbf{\Theta}$ = computeIK($\mathbf{P}_{arm}$)\;
$\mathbf{\pi_{\theta}}$ = motionPlan($\mathbf{\Theta}$)\;
executeArmMotion($\mathbf{\pi_{\theta}}$)\;

\tcp{Feedback Loop for Tracking}
\While{Arm is in motion}{
    frame, depth = captureStream()\;
    $\mathbf{P}'_{arm}$ = trackAndPos(segment, frame, depth)\;
    $\mathbf{\Theta}'$ = computeIK($\mathbf{P}'_{arm}$)\;
    adjustArmMotion($\mathbf{\Theta}'$)\;
}
\tcp{Grasping}
executeGrip($I_{gripper}$, PID)\;

\caption{Mobile grasping system pseudo-algorithm}
\label{alg:overview}
\end{algorithm}

\subsection{System Overview}
The proposed robotic system integrates vision capabilities with motion control and enhanced mobility. This system is composed of two principal modules: the \textbf{Visual Interpretation Module (VIM)} and the\textbf{ Motion Control Module (MCM)}, as Fig.~\ref{fig:robot-arm-diagram} shows.
These modules are materially supported by two integral devices: a 6 degree-of-freedom (DoF) robotic arm with an end effector, and a versatile  mobile platform (see details in Sec.~\ref{subsec:platform-overview}). 
Alg.~\ref{alg:overview} shows a comprehensive algorithm of the grasping procedure.


\subsection{Visual Interpretation Module (VIM)}
The \textbf{VIM} serves as the system's sensory and interpretative cornerstone. It encompasses a depth-sensing camera and a suite of vision foundation models, with SAM and its variants.

\subsubsection{Visual Foundation Model Integration}
Central to the VIM is the integration of SAM-like visual foundation models (see the ``Object Segmentation'' stage in Alg.~\ref{alg:overview})
%
User interaction is facilitated through diverse modalities: point-and-clicks, region drawings, or spoken commands. Reacting to these cues, the \textbf{VIM} identifies and showcases the target object, overlaying its segmented outline on the live video feed, as shown in Fig.~\ref{fig:robot-arm-diagram}.

\subsubsection{Depth Estimation and Object Positioning}

After obtaining user confirmation, the \textbf{VIM} derives the three-dimensional coordinates of the designated object (see the ``Depth Measurement and Positioning'' stage in Alg.~\ref{alg:overview}).
This crucial data is then dispatched to the Motion Control Module, as shown in Fig.~\ref{fig:robot-arm-diagram}.



    
    
    



\subsection{Motion Control Module (MCM)}
\label{subsec:motion-control}
The \textbf{MCM} acts as the robotic system's action executor, translating visual cues into precise movements and ensuring effective object grasping, as shown in Fig.~\ref{fig:robot-arm-diagram} and Alg.~\ref{alg:overview}.

\subsubsection{Platform Mobility}
One advantage of our system is its mobile capability to relocate when an object is beyond the robotic arm's reach.
Specifically, the algorithm dictates that if the target's 3D coordinate $\mathbf{P}_{arm}$ is outside the arm's operational range. Such a strategy ensures that the object is always accessible, providing a foundational prerequisite for subsequent grasping actions. This mobility is facilitated by an object tracking-based navigation system, allowing for dynamic adjustments based on the real-time location of the target.

\subsubsection{Motion Planning}

Central to the robot's grasping action is its ability to plan and execute motions. This involves a two-step process: the calculation of desired joint angles through inverse kinematics and then determining the arm's trajectory for movement.
As described in Alg.~\ref{alg:overview}, where $\mathbf{\Theta}$ represents the desired joint angles for the robotic arm, and $\mathbf{\pi_{\theta}}$ illustrates the planned movement trajectory. The process ensures that the end effector can effectively approach the target object in an optimized and collision-free manner.
The arm's kinematic behavior is modeled mathematically using the Denavit-Hartenberg (D-H) representation \cite{denavit1955kinematic}, ensuring a systematic approach to finding its potential configurations.

\subsubsection{Tracking Feedback}

Maintaining the arm's trajectory accuracy is paramount, especially when accounting for potential errors arising from depth perception or servo movements.
This continuous adjustment mechanism ensures the system remains aligned with the target even amidst unforeseen discrepancies.

\subsubsection{Grasping Mechanism}
Once the arm is in the desired position, the final step is the actual act of grasping.
The system utilizes a grip force modulated by a PID controller\cite{visioli2006practical}, which ensures adaptability across objects of varying sizes, shapes, and fragility. The term $I_{gripper}$ represents the current or force reading from the gripper, facilitating feedback-driven grip adjustments.

\label{sec:motion-planning}

\begin{figure}[tb]
    \centering
    \begin{subfigure}[t]{0.46\linewidth}
        \centering
         \includegraphics[width=\textwidth]{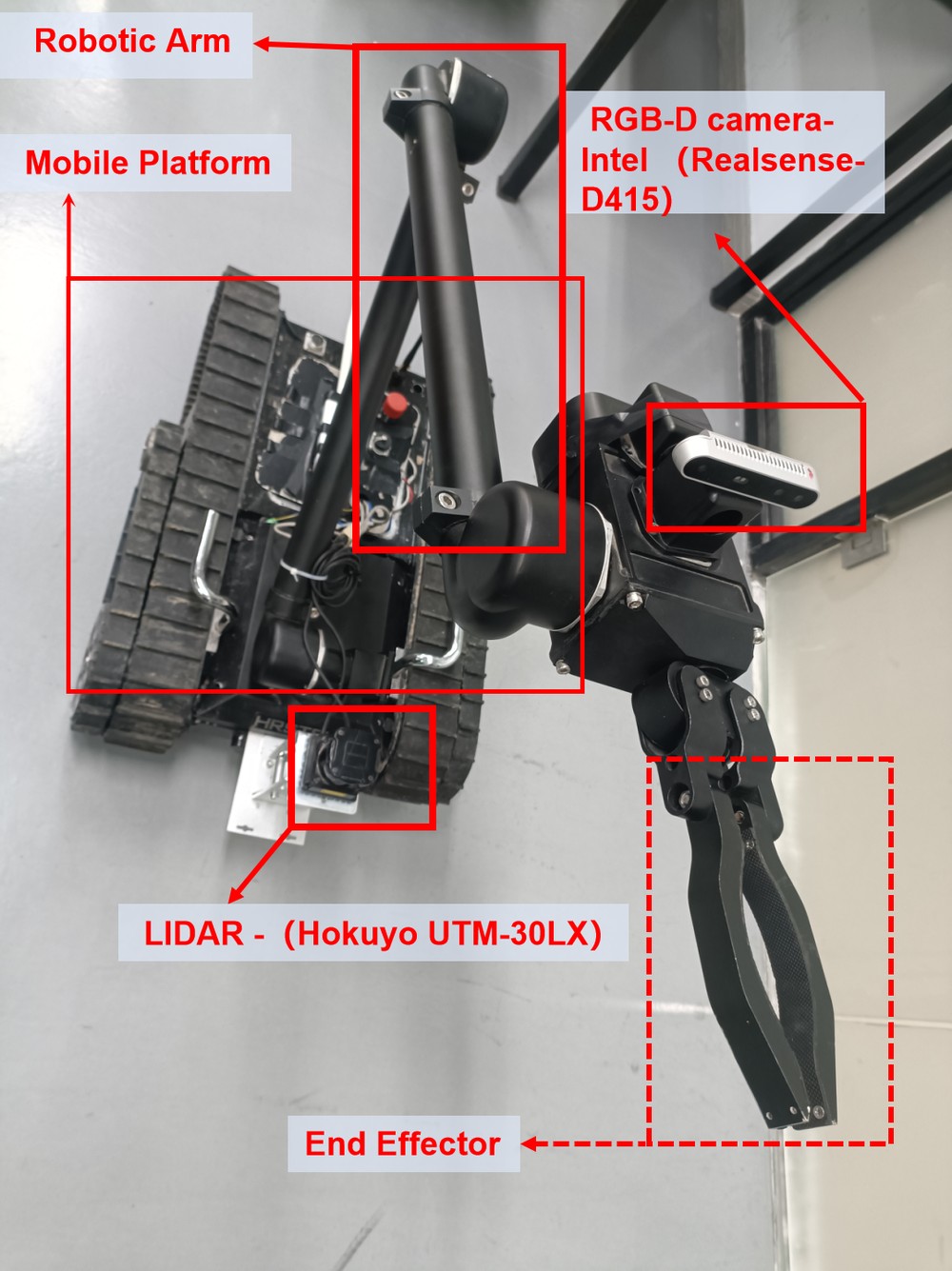}
         \caption{\small \label{subfig:system_annotated_1}}
    \end{subfigure}
    \hfill
    \begin{subfigure}[t]{0.46\linewidth}
        \centering
         \includegraphics[width=\textwidth]{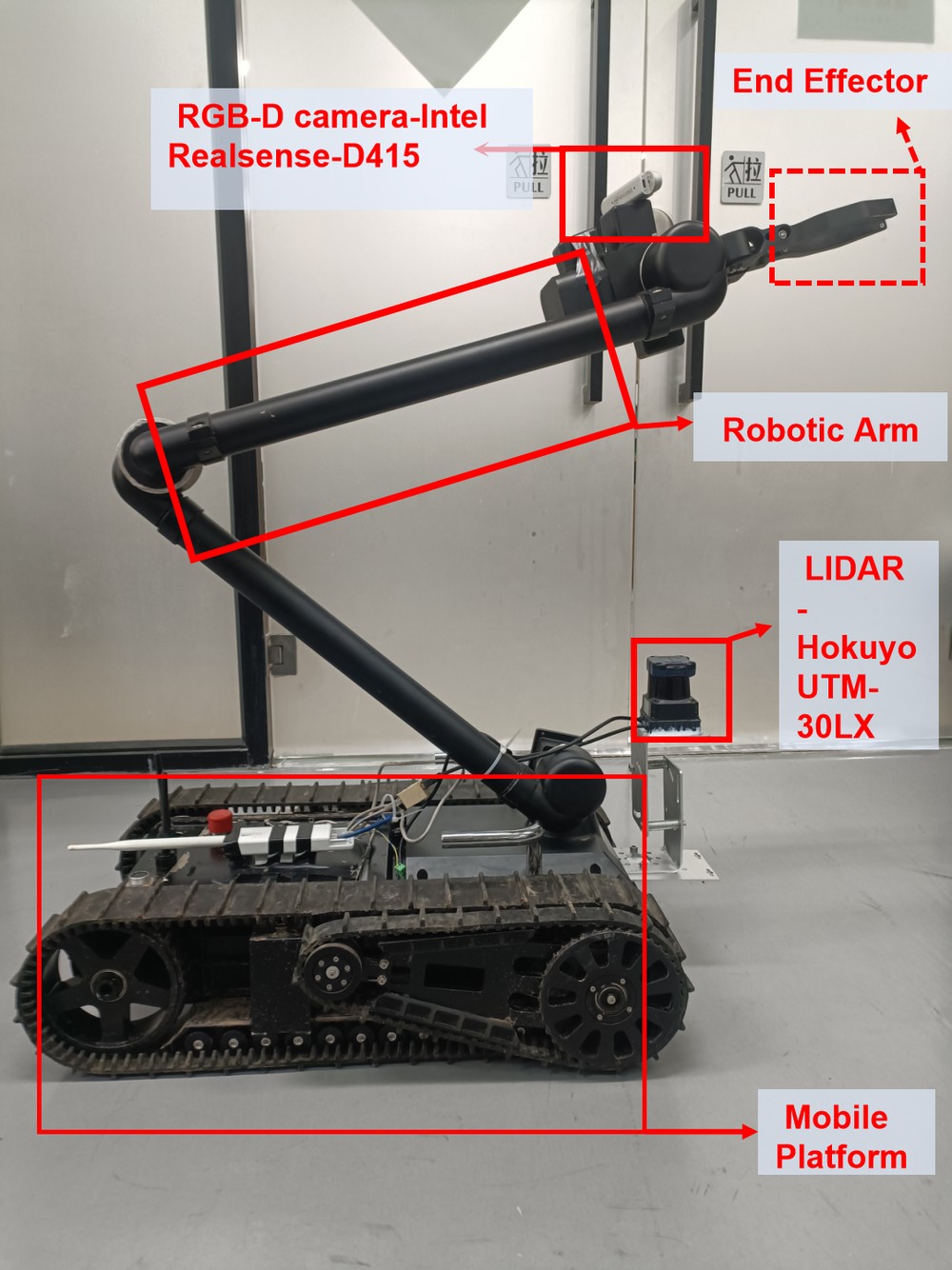}
         \caption{\small \label{subfig:system_annotated_2}}
    \end{subfigure}
    
    \caption{Mobile Grasping Platform overview from different viewpoints}
    \label{fig:system-overview}
\end{figure}

\begin{figure*}[tb]
    \centering
    \begin{subfigure}[t]{0.24\linewidth}
        \centering
         \includegraphics[width=\textwidth]{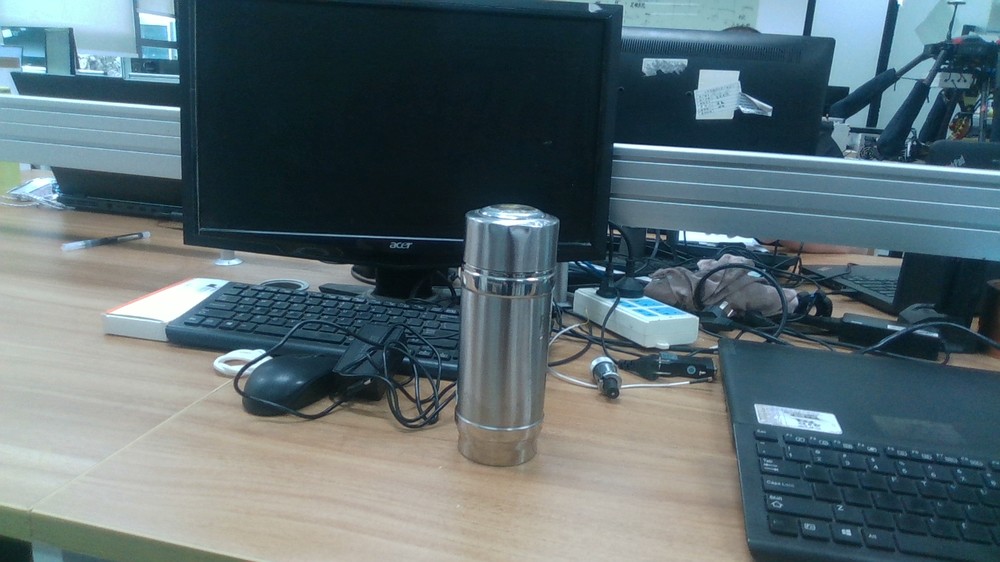}
    \end{subfigure}
    \hfill
    \begin{subfigure}[t]{0.24\linewidth}
        \centering
         \includegraphics[width=\textwidth]{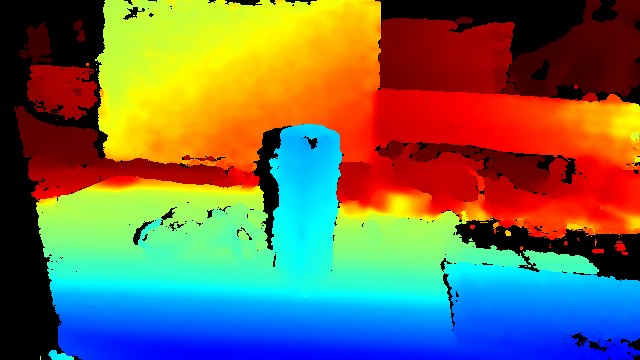}
    \end{subfigure}
    \hfill
    \begin{subfigure}[t]{0.24\linewidth}
        \centering
         \includegraphics[width=\textwidth]{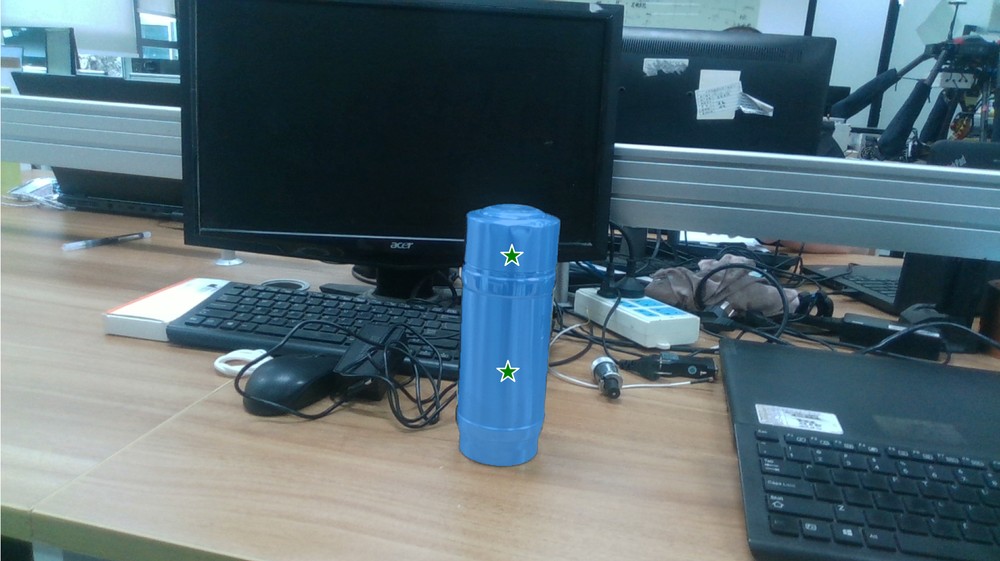}
    \end{subfigure}
    \hfill
    \begin{subfigure}[t]{0.24\linewidth}
        \centering
         \includegraphics[width=\textwidth]{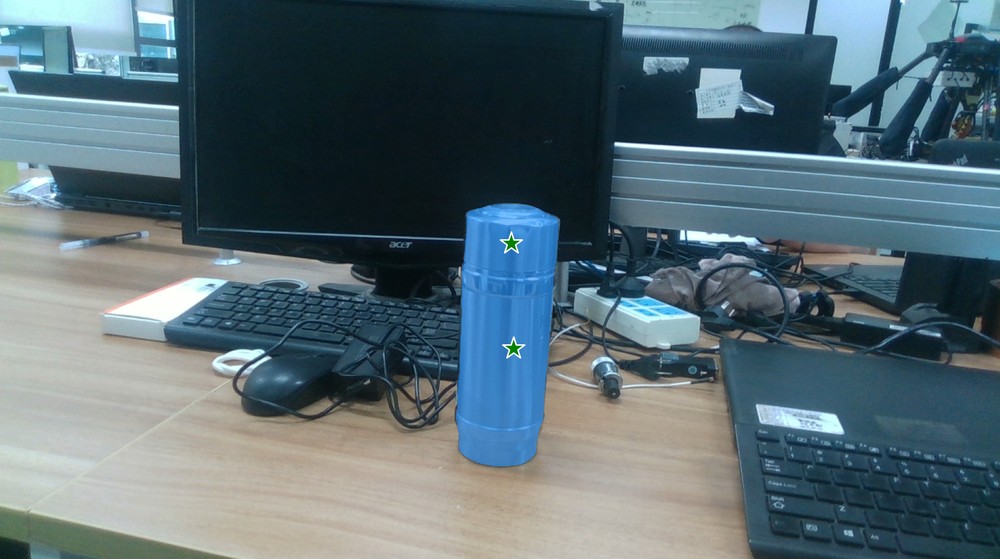}
    \end{subfigure}
    
    \vspace{1em}
    \begin{subfigure}[t]{0.24\linewidth}
        \centering
         \includegraphics[width=\textwidth]{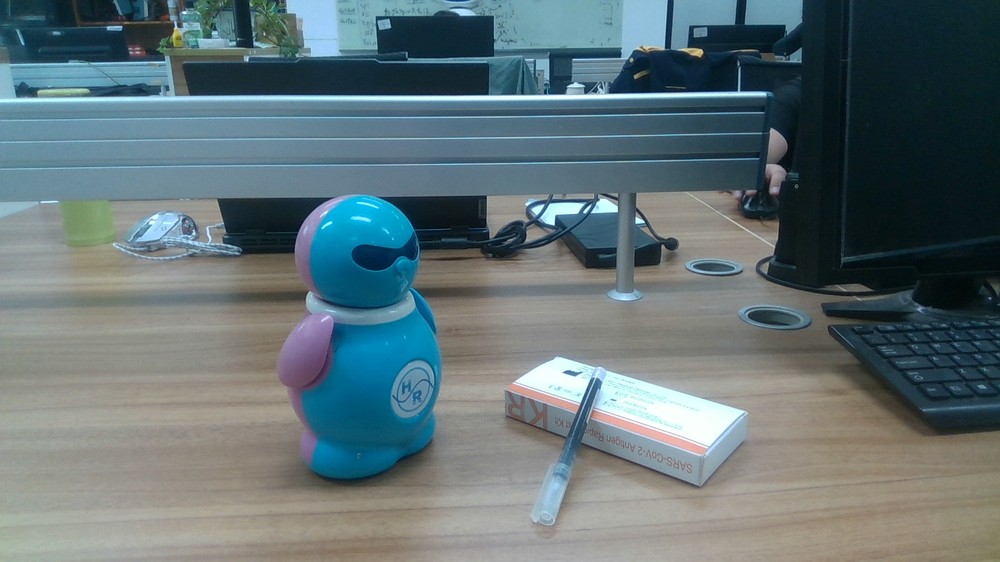}
    \end{subfigure}
    \hfill
    \begin{subfigure}[t]{0.24\linewidth}
        \centering
         \includegraphics[width=\textwidth]{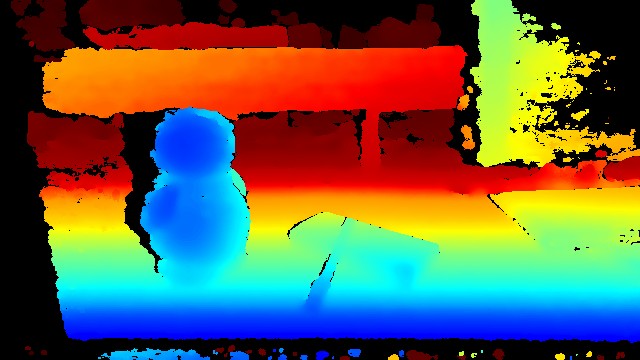}
    \end{subfigure}
    \hfill
    \begin{subfigure}[t]{0.24\linewidth}
        \centering
         \includegraphics[width=\textwidth]{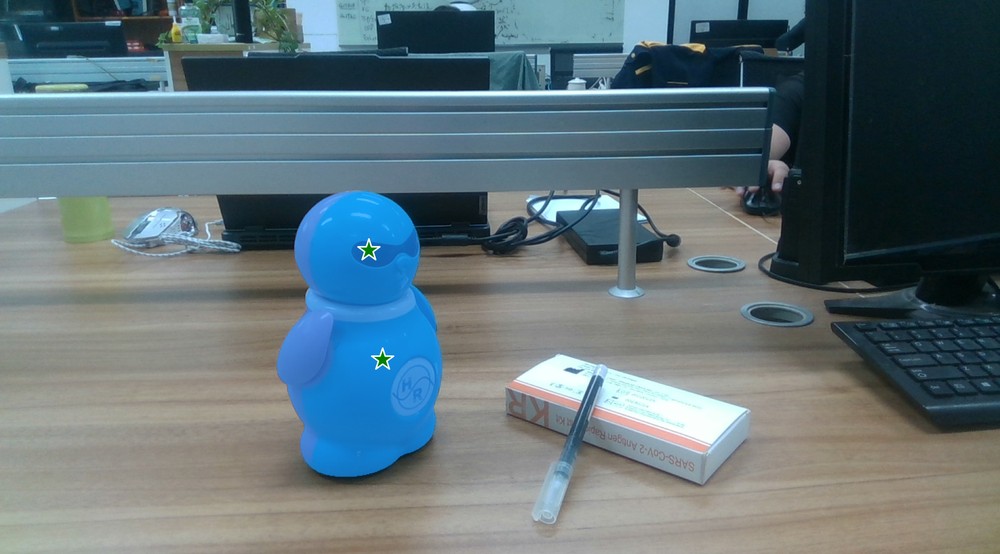}
    \end{subfigure}
    \hfill
    \begin{subfigure}[t]{0.24\linewidth}
        \centering
         \includegraphics[width=\textwidth]{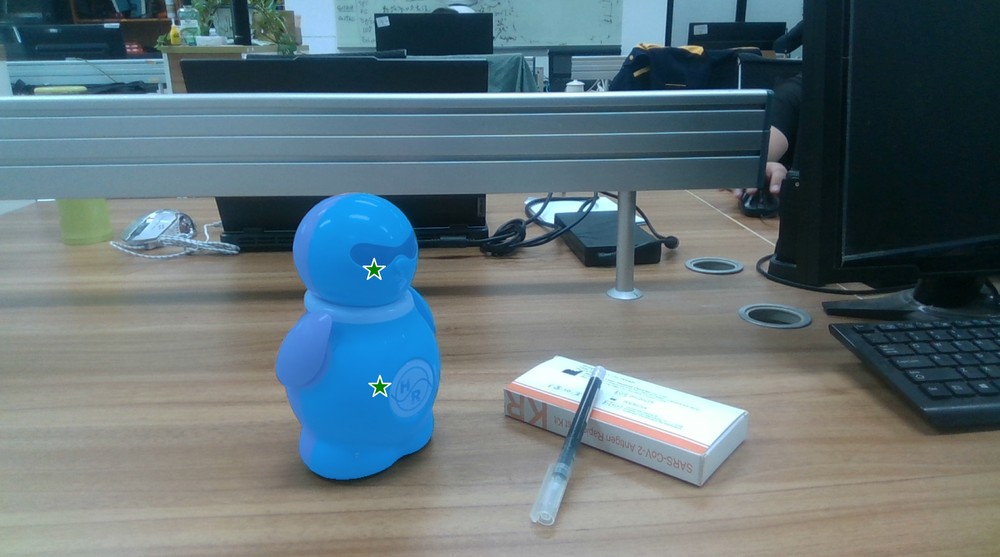}
    \end{subfigure}
    
    \vspace{1em}
    \begin{subfigure}[t]{0.24\linewidth}
        \centering
         \includegraphics[width=\textwidth]{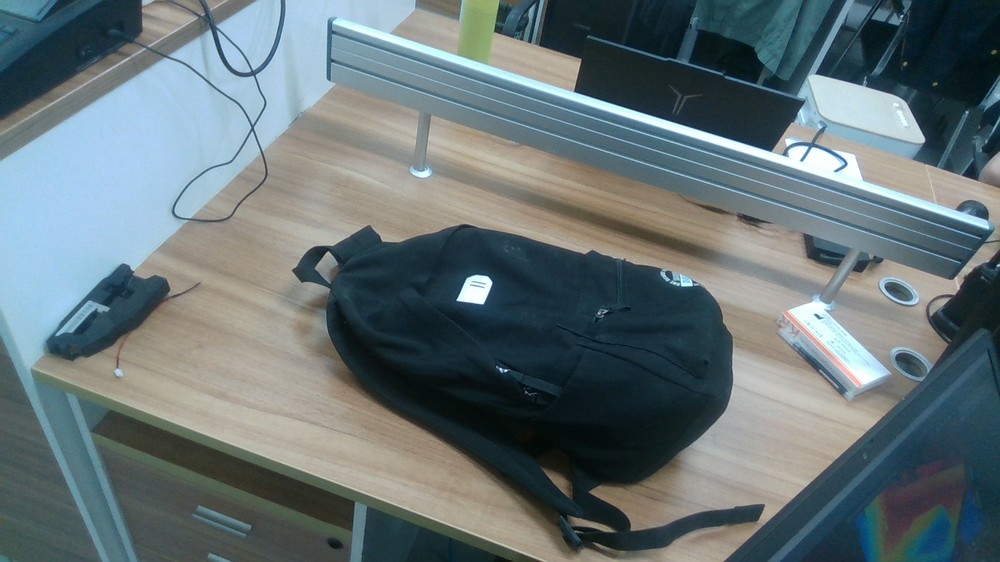}
    \end{subfigure}
    \hfill
    \begin{subfigure}[t]{0.24\linewidth}
        \centering
         \includegraphics[width=\textwidth]{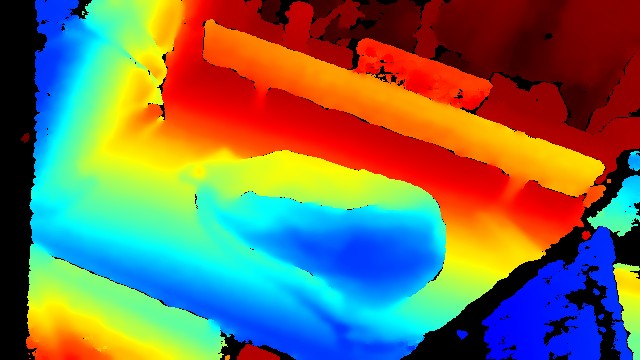}
    \end{subfigure}
    \hfill
    \begin{subfigure}[t]{0.24\linewidth}
        \centering
         \includegraphics[width=\textwidth]{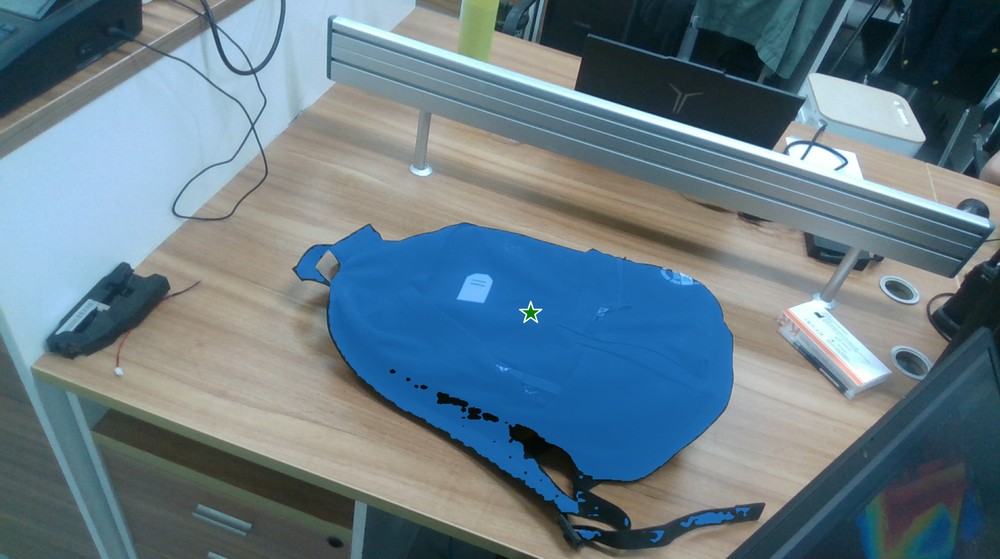}
    \end{subfigure}
    \hfill
    \begin{subfigure}[t]{0.24\linewidth}
        \centering
         \includegraphics[width=\textwidth]{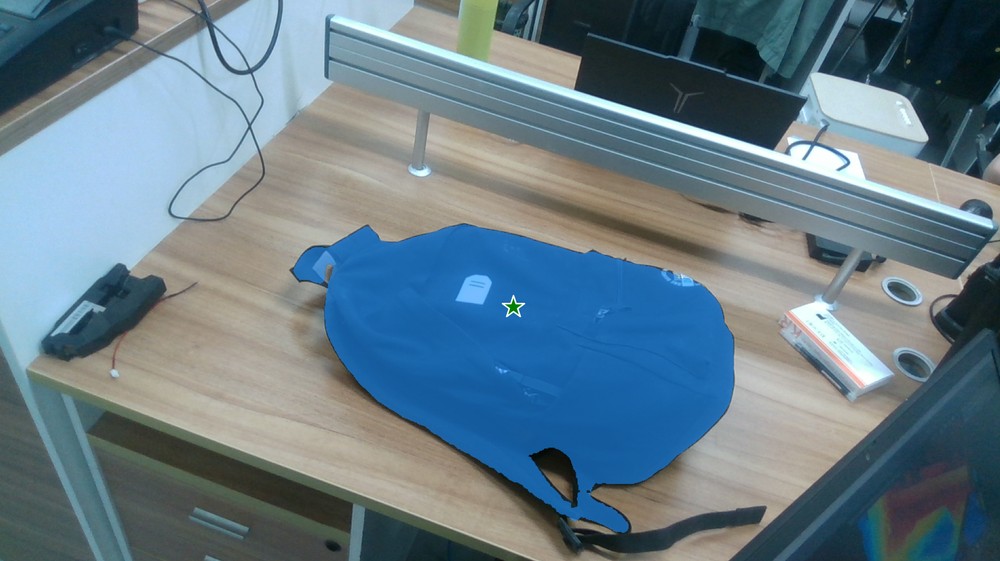}
    \end{subfigure}

    \vspace{1em}

    \begin{subfigure}[t]{0.24\linewidth}
        \centering
         \includegraphics[width=\textwidth]{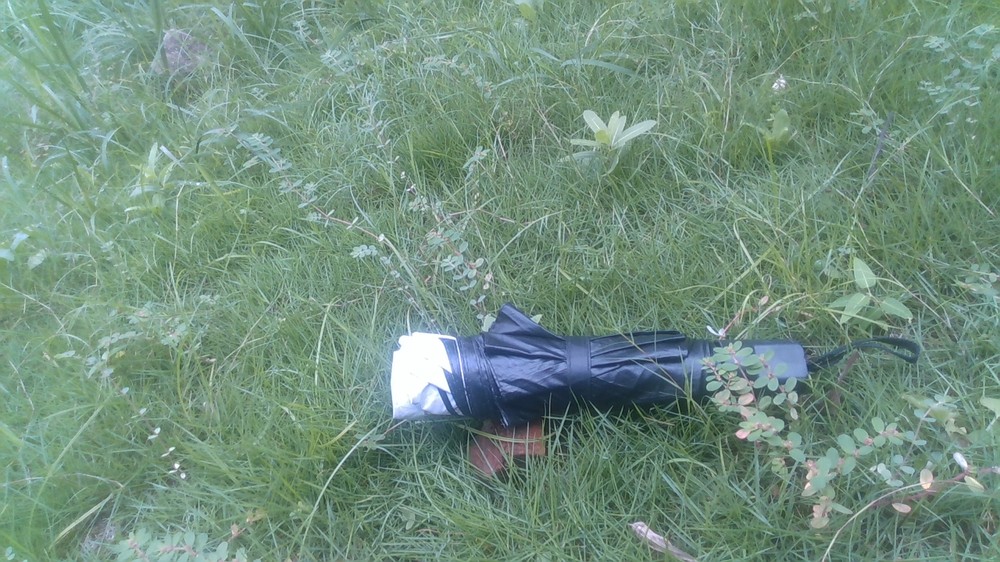}
    \end{subfigure}
    \hfill
    \begin{subfigure}[t]{0.24\linewidth}
        \centering
         \includegraphics[width=\textwidth]{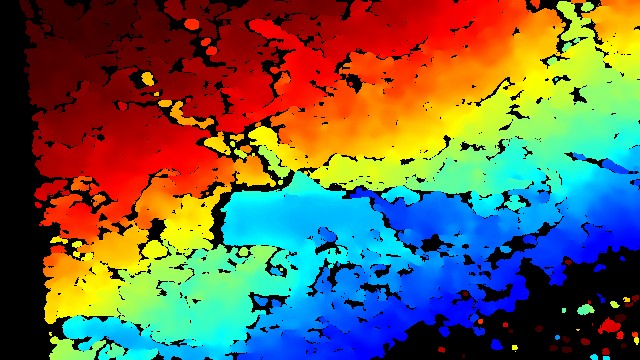}
    \end{subfigure}
    \hfill
    \begin{subfigure}[t]{0.24\linewidth}
        \centering
         \includegraphics[width=\textwidth]{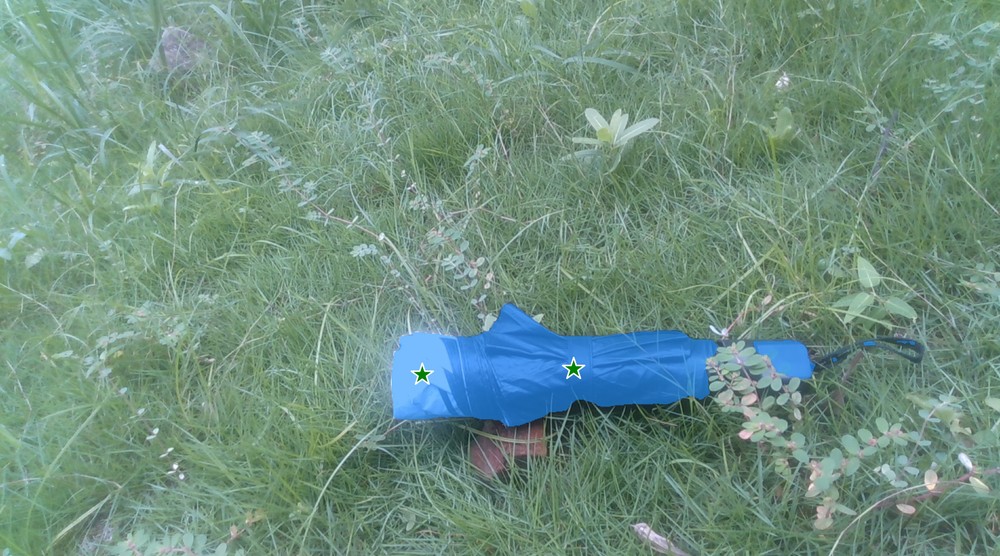}
    \end{subfigure}
    \hfill
    \begin{subfigure}[t]{0.24\linewidth}
        \centering
         \includegraphics[width=\textwidth]{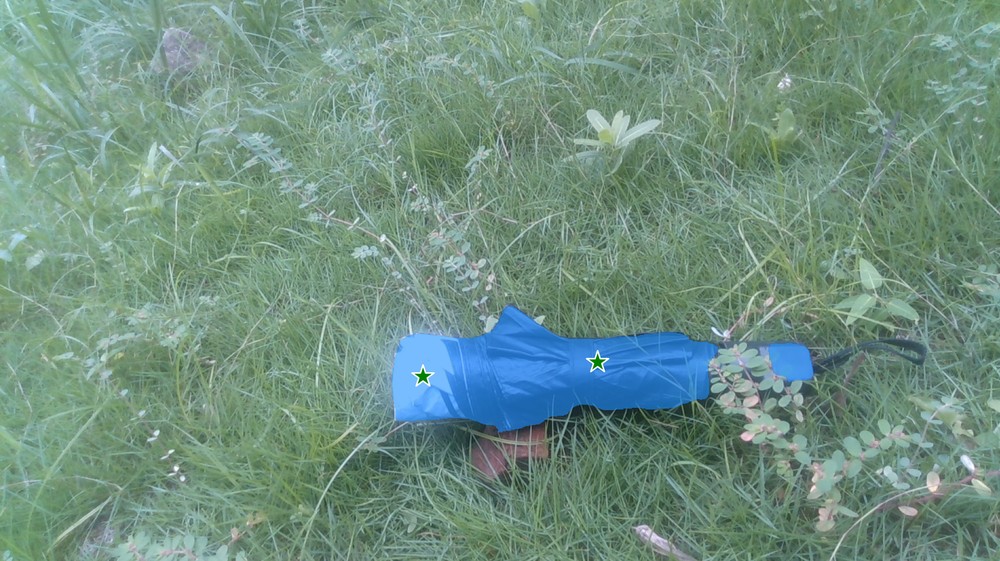}
    \end{subfigure}

    \vspace{1em}
    
    \begin{subfigure}[t]{0.24\linewidth}
        \centering
         \includegraphics[width=\textwidth]{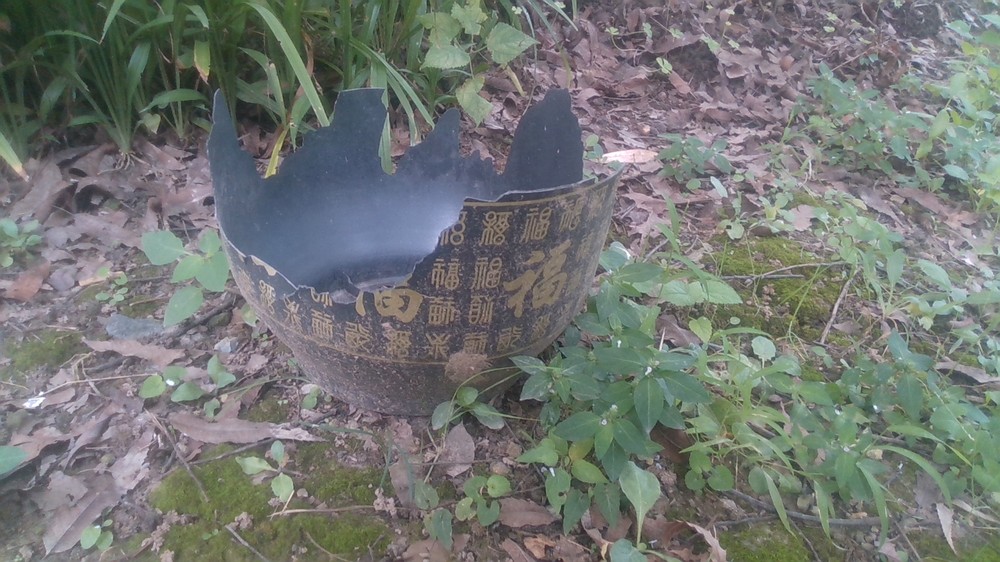}
         \caption{\small\label{subfig:sam_original} Raw Image}
    \end{subfigure}
    \hfill
    \begin{subfigure}[t]{0.24\linewidth}
        \centering
         \includegraphics[width=\textwidth]{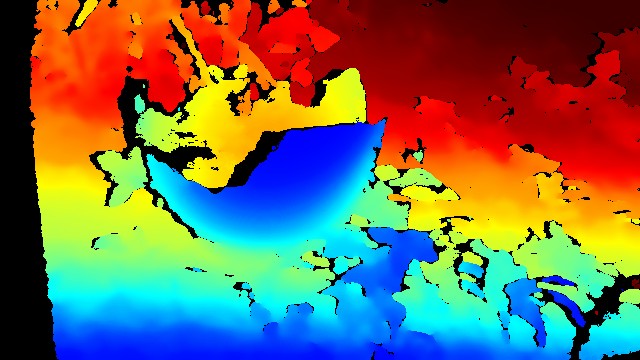}
         \caption{\small\label{subfig:depth} Depth Map}
    \end{subfigure}
    \hfill
    \begin{subfigure}[t]{0.24\linewidth}
        \centering
         \includegraphics[width=\textwidth]{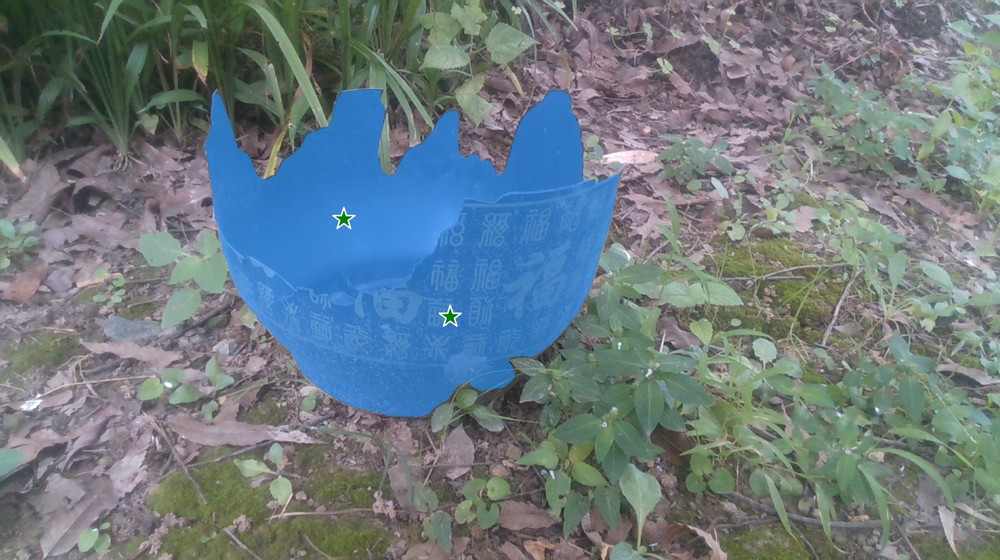}
         \caption{\small\label{subfig:sam_segment} SAM \cite{kirillov2023segment}}
    \end{subfigure}
    \hfill
    \begin{subfigure}[t]{0.24\linewidth}
        \centering
         \includegraphics[width=\textwidth]{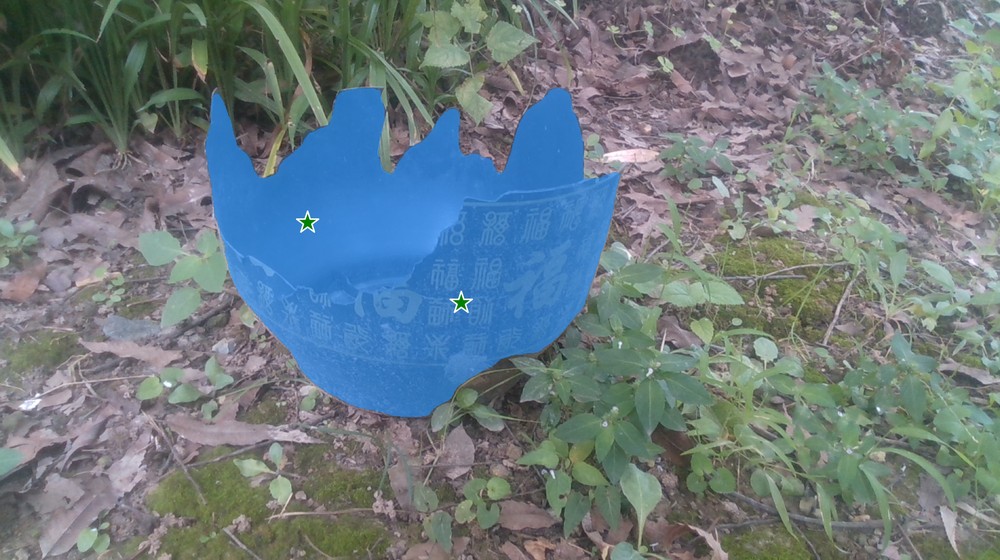}
         \caption{\small\label{subfig:mobile_segment} Mobile SAM \cite{zhang2023faster}}
    \end{subfigure}
    
    \caption{Comparison of SAM \cite{kirillov2023segment} and Mobile-SAM \cite{zhang2023faster} performance on both indoor and outdoor scenarios. 
    Zoom-in for the best view.
    }
    \label{fig:sam_mobile_comparison}
\end{figure*}

\section{Experiments and Results}
\label{sec:results}

\subsection{Platform Configuration Overview}
\label{subsec:platform-overview}
Our platform is a comprehensive assembly of a 6-DoF robotic arm and a mobile platform, specifically an all-terrain crawler AGV, as shown in Fig.~\ref{fig:system-overview} shows.

\subsubsection{Camera Selection}
For depth perception, our experimental setup employed the Intel Realsense-D415 depth camera. Positioned at the pinnacle of the arm, adjacent to the end-effector, this camera supports the ``eye-in-hand'' grasping approach.

\subsubsection{Robotic Arm Joints}
Our robotic arm, with its six degrees of freedom, is articulated through:
\begin{itemize}
    \item \textbf{Joint-0}: Base Horizontal Rotation 
    \item \textbf{Joint-1}: Upper Arm Pitch 
    \item \textbf{Joint-2}: Elbow Pitch 
    \item \textbf{Joint-3}: Wrist Pitch 
    \item \textbf{Joint-4}: Wrist Radial Rotation
    \item \textbf{Joint-5}: Gripper Open/Close 
\end{itemize}

\subsubsection{Mobile Platform Configuration}
In our experiment, the mobile platform employs the Hokuyo UTM-30LX LIDAR. This choice complements the depth camera by complimenting its view blind spots, supporting a more holistic sensing of the environment. Certain SLAM and navigation algorithms \cite{zhang2019hierarchical, macenski2021slam} are employed to support the movements.


\subsection{Performance Comparison: Original SAM vs. Mobile SAM}
\label{subsec:sam_mobile_comparison}
For efficient, lightweight visual interpretation, we conducted a comparative analysis between the original SAM and Mobile SAM. 
Demonstrative examples are presented in Fig.~\ref{fig:sam_mobile_comparison}.
Across diverse environments (both indoor and outdoor), Mobile SAM matches the performance of the original SAM given identical prompts. Notably, Mobile SAM's model size is 60 times smaller than its predecessor, facilitating deployment on standard home-grade graphics cards. This means it's well-suited for economical industrial computers with GPU support. 
Based on our tests, Mobile SAM can consistently deliver rapid responses in around 50ms, with a single NVIDIA 3060 graphics card.

Fig.~\ref{fig:sam_mobile_comparison} also showcases various applications for robotic grasping. The module adeptly segments a spectrum of objects, from daily items to more specialized targets such as suspicious packages removal.


\subsection{Simulated Grasping Results}

\begin{figure}
    \centering
    \begin{subfigure}[t]{0.31\linewidth}
        \centering
         \includegraphics[width=\textwidth]{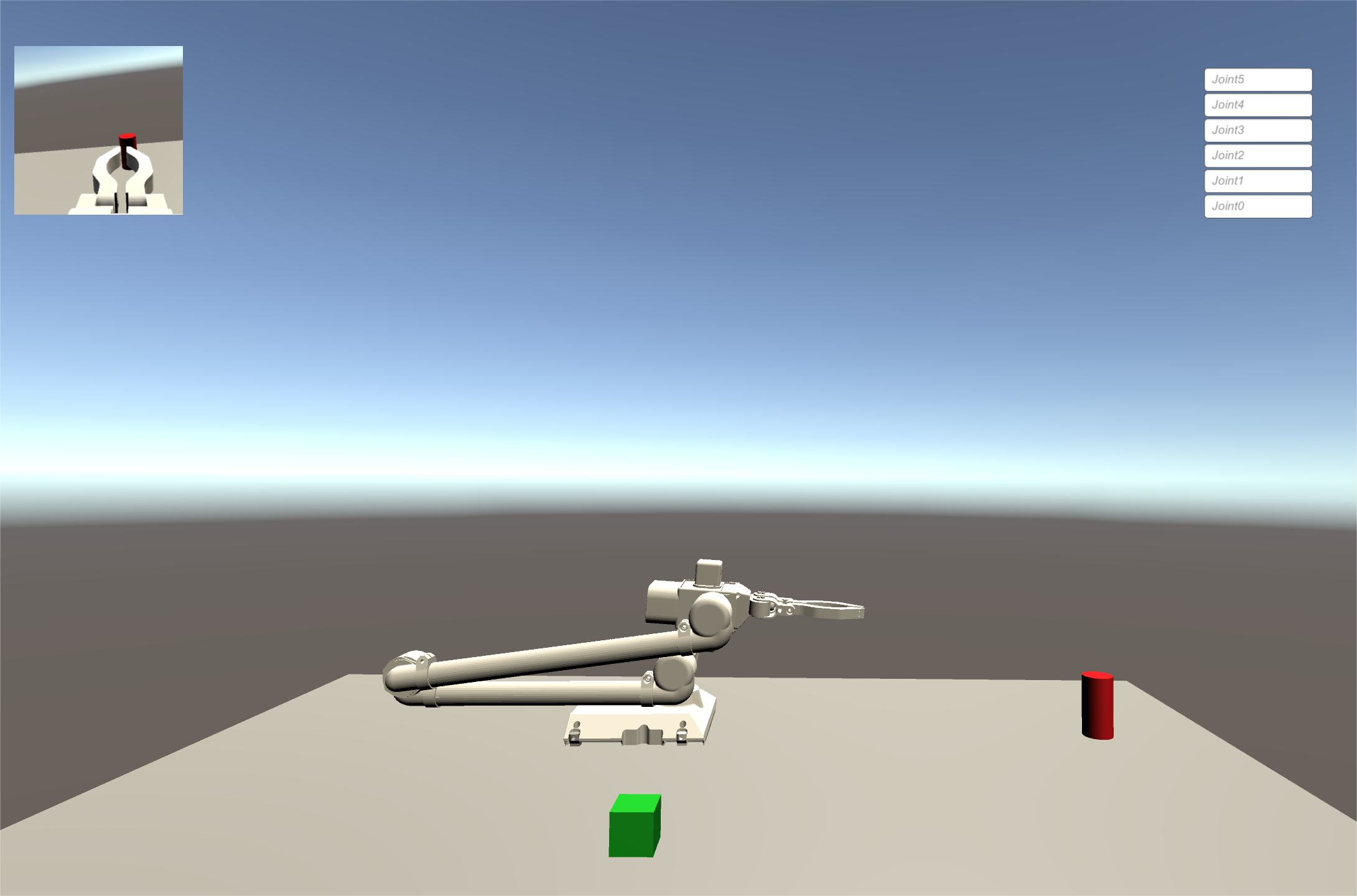}
         \caption{\small Initial pose}
    \end{subfigure}
    \hfill
    \begin{subfigure}[t]{0.31\linewidth}
        \centering
         \includegraphics[width=\textwidth]{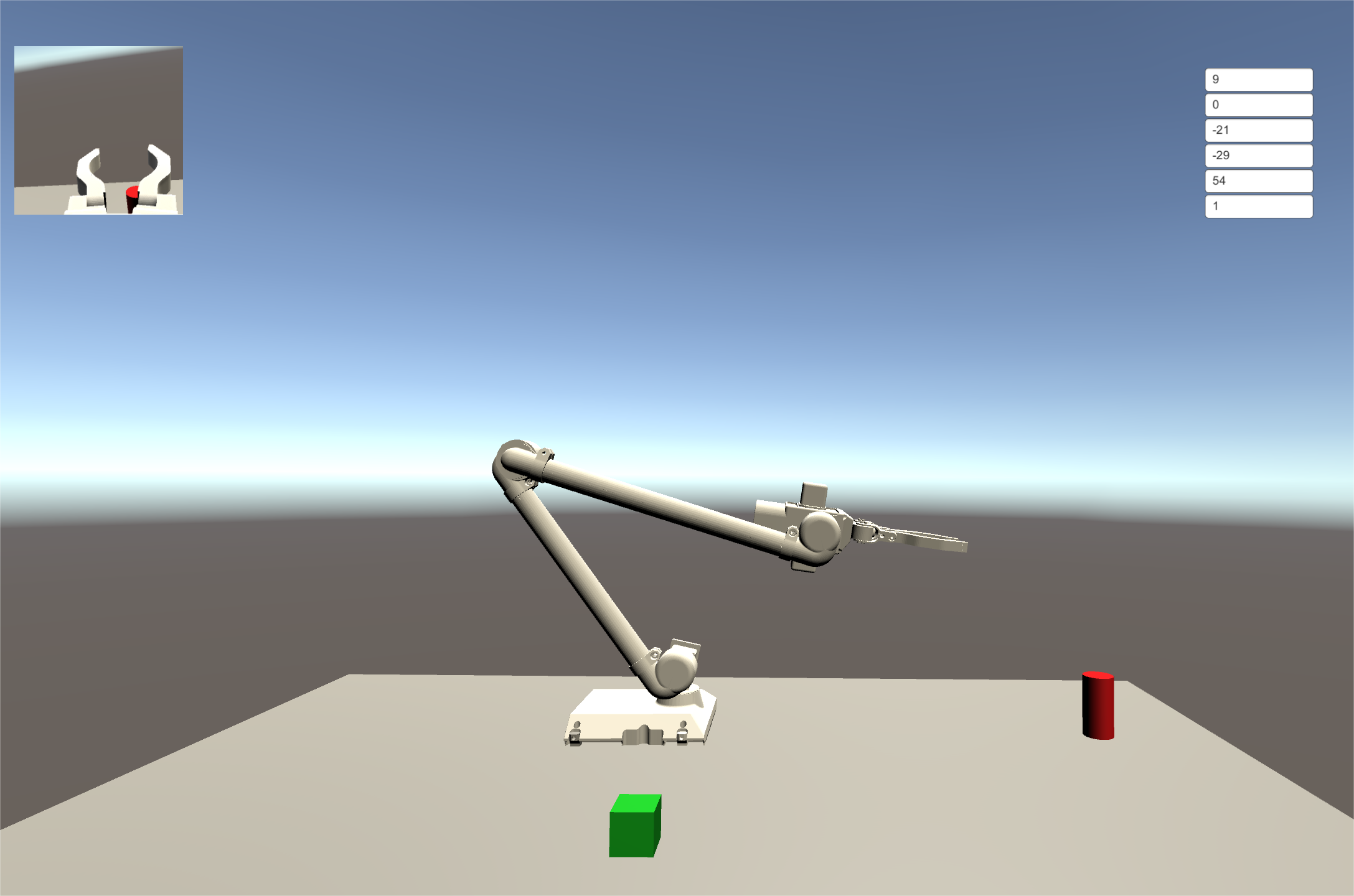}
         \caption{\small Midway pose}
    \end{subfigure}
    \hfill
    \begin{subfigure}[t]{0.31\linewidth}
        \centering
         \includegraphics[width=\textwidth]{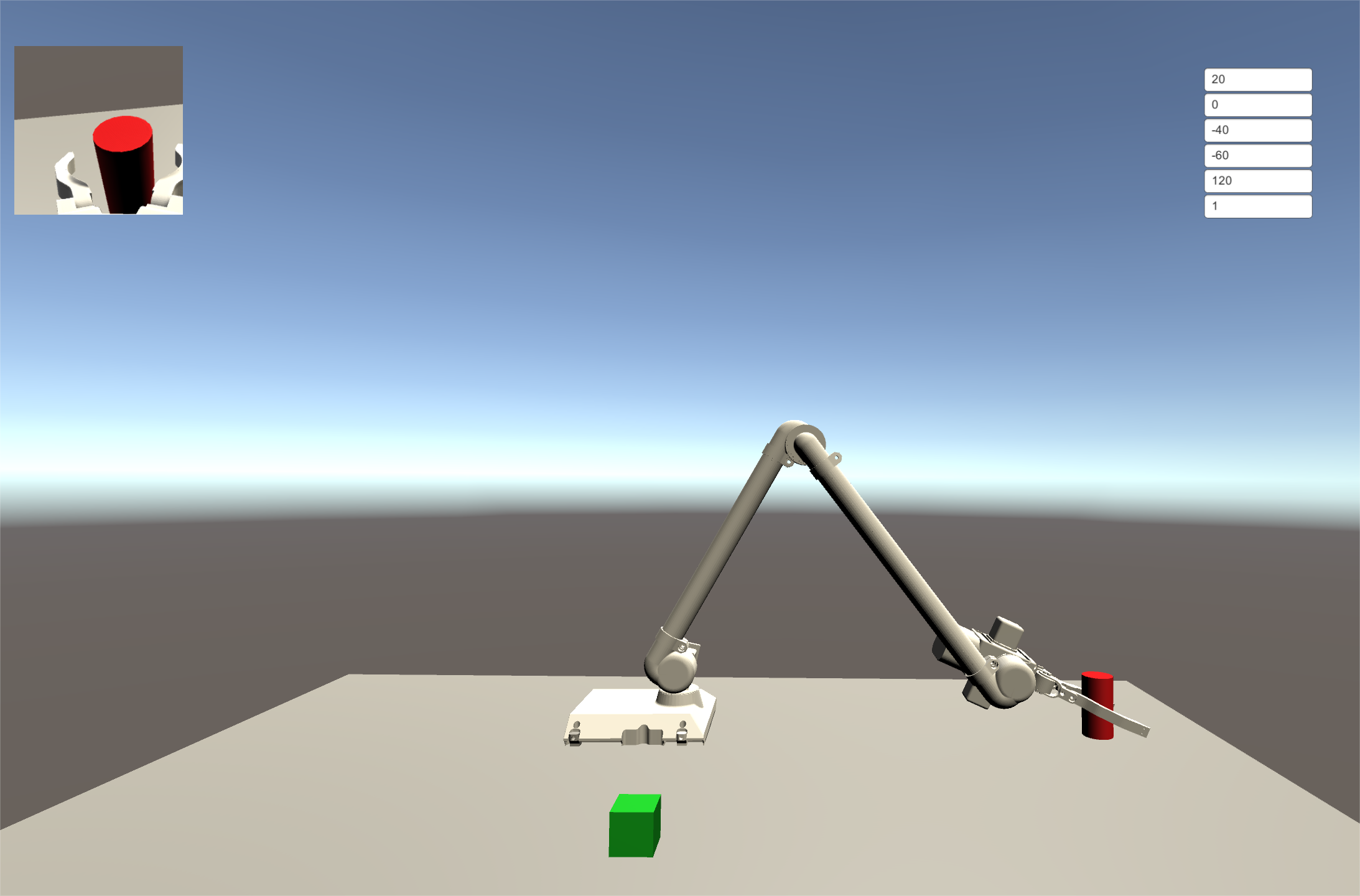}
         \caption{\small Final pose}
    \end{subfigure}
    \caption{The simulated grasping process is delineated across three primary phases: initial, midway, and final poses. The analogue camera's viewpoint is presented on the top-left, while corresponding joint angles are displayed on the top-right. 
    Zoom-in for the best view.
    }
    \label{fig:simulation}
\end{figure}

Our preliminary testing process began with simulating the grasping sequence, depicted in Fig.~\ref{fig:simulation}. In this simulation, an analogue camera captures the object's initial state. This video feed is subsequently relayed to our visual interpretation module, facilitating segmentation and derivation of the 3D object coordinates for the grasping action. Desired joint angles are then determined using our motion control module, which leverages inverse kinematic estimation coupled with the robotic arm's D-H model configuration. To ensure safety and assess feasibility, joint movements are simulated before execution. Importantly, our actual robotic arm is interfaced with this simulation, enabling synchronized movements between the two. This simulation framework incorporates both our robotic arm model and the depth camera, highlighting the advantages of our integrated eye-in-hand system.

\subsection{Real-World Grasping Results}

\begin{figure}
    \centering
    \begin{subfigure}[t]{0.46\linewidth}
        \centering
         \includegraphics[width=\textwidth]{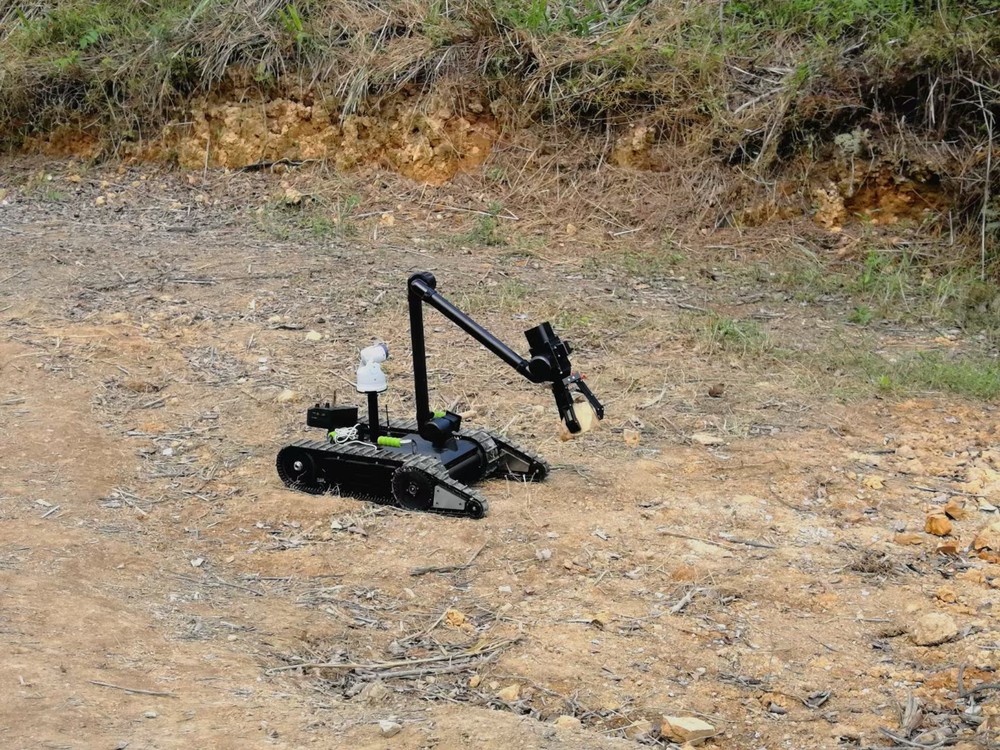}
         \caption{\small \label{subfig:grasp_demo_1}}
    \end{subfigure}
    \hfill
    \begin{subfigure}[t]{0.46\linewidth}
        \centering
         \includegraphics[width=\textwidth]{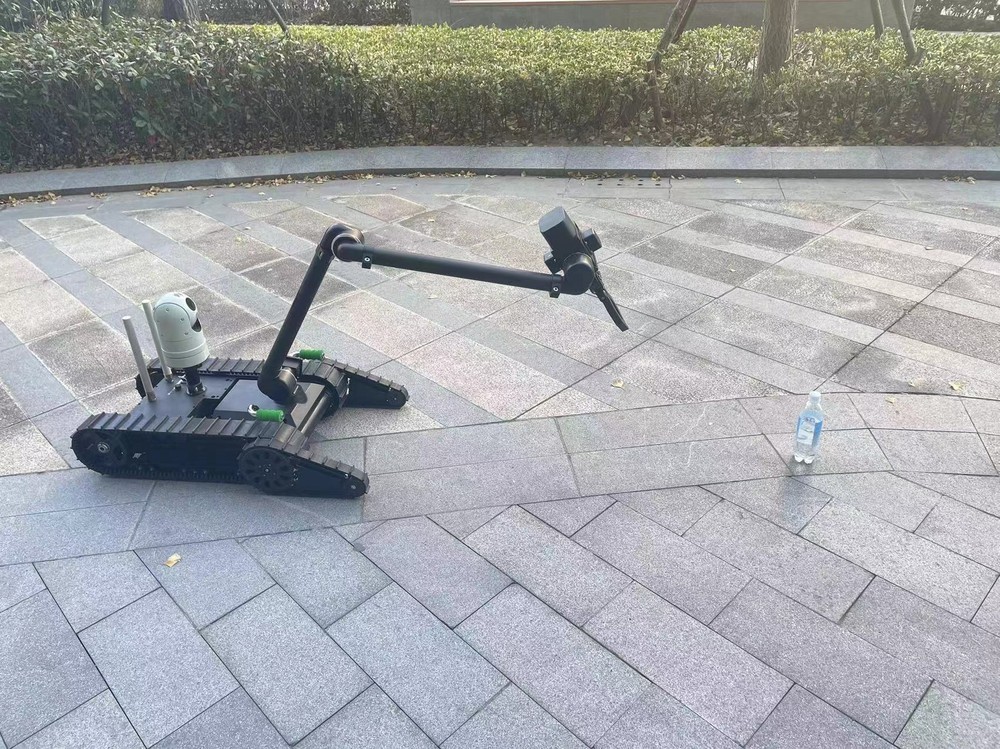}
         \caption{\small \label{subfig:grasp_demo_2}}
    \end{subfigure}

    \caption{Our mobile grasping platform performs grasping in different scenarios.}
    \label{fig:grasp_demo}
\end{figure}

Progressing from simulation, we transitioned our algorithms to the actual robotic system. A salient feature of our system, afforded by the mobile platform, is its ability to bridge spatial gaps. If a target is beyond the arm's immediate reach, the robot autonomously navigates closer, guided by object tracking. Once close, it executes the grasping procedure. 
%
%
Importantly, the system demonstrated its versatility by operating proficiently in both indoor and outdoor settings, as Fig.~\ref{fig:grasp_demo} demonstrates.


\section{Conclusion \& Future Work}
\label{sec:conclusion}

The integration of visual foundation model with robotic arms placed on mobile platforms represents a new stage in the field of robotics. This combination ensures a more intuitive user experience, as individuals can effortlessly guide the robot by clicking, drawing, or speaking, directing it towards the desired object for grasping. The system's adaptability allows it to segment and grasp a wide variety of items, expanding its applicability across numerous domains.
Furthermore, leveraging the visual foundation model as a "zero-shot" detector, our system avoids the traditional, tedious training and testing cycles associated with grasping robotic arms in traditional industrial settings.




In terms of future prospects and enhancements, we aim to further refine the grasping algorithm by incorporating the detailed contours provided by the visual foundation models. Understanding an object's contour can not only improve the grasping gesture but also minimize any potential damage to sensitive items. This becomes especially crucial in delicate operations, such as grasping a fragile cup for disabled people, where any deformation might cause unintended consequences.

While the mobile SAM variant offers a significant reduction in size and increased processing speed compared to its predecessor, visual foundation models remain computationally demanding. Our ongoing ambition is to achieve real-time segmentation with diminished GPU reliance, possibly even eliminating it. Approaches such as knowledge distillation and model quantification techniques stand as promising avenues to achieve this goal.






{\small
\bibliographystyle{ieee_fullname}
\bibliography{reference}

\begin{thebibliography}{10}\itemsep=-1pt

\bibitem{bang2023multitask}
Yejin Bang, Samuel Cahyawijaya, Nayeon Lee, Wenliang Dai, Dan Su, Bryan Wilie, Holy Lovenia, Ziwei Ji, Tiezheng Yu, Willy Chung, et~al.
\newblock A multitask, multilingual, multimodal evaluation of chatgpt on reasoning, hallucination, and interactivity.
\newblock {\em arXiv preprint arXiv:2302.04023}, 2023.

\bibitem{bubeck2023sparks}
S{\'e}bastien Bubeck, Varun Chandrasekaran, Ronen Eldan, Johannes Gehrke, Eric Horvitz, Ece Kamar, Peter Lee, Yin~Tat Lee, Yuanzhi Li, Scott Lundberg, et~al.
\newblock Sparks of artificial general intelligence: Early experiments with gpt-4.
\newblock {\em arXiv preprint arXiv:2303.12712}, 2023.

\bibitem{chen2023bridging}
Zhimin Chen and Bing Li.
\newblock Bridging the domain gap: Self-supervised 3d scene understanding with foundation models.
\newblock {\em arXiv preprint arXiv:2305.08776}, 2023.

\bibitem{cheng2023segment}
Yangming Cheng, Liulei Li, Yuanyou Xu, Xiaodi Li, Zongxin Yang, Wenguan Wang, and Yi Yang.
\newblock Segment and track anything.
\newblock {\em arXiv preprint arXiv:2305.06558}, 2023.

\bibitem{denavit1955kinematic}
Jacques Denavit and Richard~S Hartenberg.
\newblock A kinematic notation for lower-pair mechanisms based on matrices.
\newblock 1955.

\bibitem{du2021vision}
Guoguang Du, Kai Wang, Shiguo Lian, and Kaiyong Zhao.
\newblock Vision-based robotic grasping from object localization, object pose estimation to grasp estimation for parallel grippers: a review.
\newblock {\em Artificial Intelligence Review}, 54(3):1677--1734, 2021.

\bibitem{fan2023pope}
Zhiwen Fan, Panwang Pan, Peihao Wang, Yifan Jiang, Dejia Xu, Hanwen Jiang, and Zhangyang Wang.
\newblock Pope: 6-dof promptable pose estimation of any object, in any scene, with one reference.
\newblock {\em arXiv preprint arXiv:2305.15727}, 2023.

\bibitem{huang2023voxposer}
Wenlong Huang, Chen Wang, Ruohan Zhang, Yunzhu Li, Jiajun Wu, and Li Fei-Fei.
\newblock Voxposer: Composable 3d value maps for robotic manipulation with language models.
\newblock {\em arXiv preprint arXiv:2307.05973}, 2023.

\bibitem{jiang2022review}
Peiyuan Jiang, Daji Ergu, Fangyao Liu, Ying Cai, and Bo Ma.
\newblock A review of yolo algorithm developments.
\newblock {\em Procedia Computer Science}, 199:1066--1073, 2022.

\bibitem{kirillov2023segment}
Alexander Kirillov, Eric Mintun, Nikhila Ravi, Hanzi Mao, Chloe Rolland, Laura Gustafson, Tete Xiao, Spencer Whitehead, Alexander~C Berg, Wan-Yen Lo, et~al.
\newblock Segment anything.
\newblock {\em arXiv preprint arXiv:2304.02643}, 2023.

\bibitem{liu2016ssd}
Wei Liu, Dragomir Anguelov, Dumitru Erhan, Christian Szegedy, Scott Reed, Cheng-Yang Fu, and Alexander~C Berg.
\newblock Ssd: Single shot multibox detector.
\newblock In {\em Computer Vision--ECCV 2016: 14th European Conference, Amsterdam, The Netherlands, October 11--14, 2016, Proceedings, Part I 14}, pages 21--37. Springer, 2016.

\bibitem{macenski2021slam}
Steve Macenski and Ivona Jambrecic.
\newblock Slam toolbox: Slam for the dynamic world.
\newblock {\em Journal of Open Source Software}, 6(61):2783, 2021.

\bibitem{vemprala2023chatgpt}
Sai Vemprala, Rogerio Bonatti, Arthur Bucker, and Ashish Kapoor.
\newblock Chatgpt for robotics: Design principles and model abilities.
\newblock {\em Microsoft Auton. Syst. Robot. Res}, 2:20, 2023.

\bibitem{visioli2006practical}
Antonio Visioli.
\newblock {\em Practical PID control}.
\newblock Springer Science \& Business Media, 2006.

\bibitem{yang2023pave}
Jiange Yang, Wenhui Tan, Chuhao Jin, Bei Liu, Jianlong Fu, Ruihua Song, and Limin Wang.
\newblock Pave the way to grasp anything: Transferring foundation models for universal pick-place robots.
\newblock {\em arXiv preprint arXiv:2306.05716}, 2023.

\bibitem{yu2023inpaint}
Tao Yu, Runseng Feng, Ruoyu Feng, Jinming Liu, Xin Jin, Wenjun Zeng, and Zhibo Chen.
\newblock Inpaint anything: Segment anything meets image inpainting.
\newblock {\em arXiv preprint arXiv:2304.06790}, 2023.

\bibitem{zhang2023faster}
Chaoning Zhang, Dongshen Han, Yu Qiao, Jung~Uk Kim, Sung-Ho Bae, Seungkyu Lee, and Choong~Seon Hong.
\newblock Faster segment anything: Towards lightweight sam for mobile applications.
\newblock {\em arXiv preprint arXiv:2306.14289}, 2023.

\bibitem{zhang2023survey}
Chaoning Zhang, Sheng Zheng, Chenghao Li, Yu Qiao, Taegoo Kang, Xinru Shan, Chenshuang Zhang, Caiyan Qin, Francois Rameau, Sung-Ho Bae, et~al.
\newblock A survey on segment anything model (sam): Vision foundation model meets prompt engineering.
\newblock {\em arXiv preprint arXiv:2306.06211}, 2023.

\bibitem{zhang2023bridging}
Shimian Zhang.
\newblock Bridging intelligence and instinct: A new control paradigm for autonomous robots.
\newblock {\em arXiv preprint arXiv:2307.10690}, 2023.

\bibitem{zhang2019hierarchical}
Shimian Zhang and Qiuhong Lu.
\newblock A hierarchical searching approach for laser slam re-localization on mobile robot platform.
\newblock In {\em 2019 IEEE International Conference on Real-time Computing and Robotics (RCAR)}, pages 464--467. IEEE, 2019.

\end{thebibliography}
}

\end{document}